\documentclass[letterpaper]{article} 
\usepackage[preprint]{format/aaai2027}  

\usepackage[
    letterpaper,
    margin=1in
]{geometry}

\usepackage[hyphens]{url}  
\usepackage{graphicx} 
\usepackage{natbib} 
\usepackage{caption} 
\usepackage{booktabs}
\usepackage{amsmath}

\def\BibTeX{{\rm B\kern-.05em{\sc i\kern-.025em b}\kern-.08em
    T\kern-.1667em\lower.7ex\hbox{E}\kern-.125emX}}

\newcommand{\Tlow}{T_{\mathrm{low}}}
\newcommand{\Thigh}{T_{\mathrm{high}}}
\newcommand{\ptarget}{p_{\mathrm{target}}}

\begin{document}

\title{Threshold-Based Early Stopping of Accumulations in Neural Networks with Binary Activation }

\author {
    Quentin Luquet de Saint-Germain,
    Massil Ait Abdeslam,
    Jean Pierre David
}
\affiliations {
    Polytechnique Montréal, Department of Electrical Engineering\\
    Montréal, Canada\\
    quentin.luquet-de-saint-germain@polymtl.ca, massil.ait-abdelsalem@polymtl.ca, jean-pierre.david@polymtl.ca 
}

\maketitle
\begin{abstract}
Binary neural networks are very attractive for constrained deployment, enabling small footprint and low-power inference. For binary activations, the dot products become sign-controlled additions or subtractions, but the number of operations is unchanged. Indeed, every neuron or output channel still accumulates all of its input, even though only the sign will be retained, which is often wasteful. As the accumulation progresses, the running partial sum frequently drifts so far from zero that its final sign becomes highly predictable long before the last term is reached; every contribution evaluated after that point changes the value of the sum but not the final output activation. This paper turns this observation into a post-training early-stopping mechanism. We characterize the behavior of the running accumulations on the training dataset and use this information to predict the final sign as soon as possible. No model parameter is retrained. We count the number of operations under an idealized ordering of weights. On VGG11 applied to the CIFAR-10 dataset, the method removes $86.6\%$ of the accumulation terms of the deepest convolution for a $0.37$-point accuracy drop, and $25\%$ of the full-network arithmetic when used on the three deepest convolutions simultaneously, for a $1.36$-point drop. 


\end{abstract}
\section{Introduction}\label{sec:intro}

Deep neural networks have achieved strong performance by increasing model depth, width, and representational capacity. This progress has come with increasing arithmetic and memory requirements, making inference more difficult under constrained computational budgets. At large scale, these computational costs ultimately translate into growing data-center electricity demand \cite{iea2025energyai}. In this context, improving AI efficiency can be viewed not only as a question of model accuracy, but also as how efficiently electrical resources are converted into useful predictions.

Addressing this challenge requires progress on two complementary fronts. On the software side, inference can be made less expensive by adapting the executed computation to the current input or to a target budget, through conditional inference, pruning, and quantization \cite{teerapittayanon2016branchynet,wang2018skipnet,gao2019dynamicchannel,lecun1989optimal,han2016deepcompression,courbariaux2015binaryconnect,courbariaux2016bnn,rastegari2016xnor}. On the hardware side, efficiency improves when processors are designed around the operations exposed by these models. GPUs made large convolutional networks practical \cite{krizhevsky2012imagenet}, while FPGAs, ASICs, and domain-specific accelerators such as TPUs push specialization further \cite{jouppi2017tpu}. Among these approaches, quantization is particularly important because it changes the numerical representation of the computation itself. Binarization takes this concept to its ultimate limit by representing weights, activations, or both using a single bit. This reduces storage and simplifies elementary operations, but it does not necessarily reduce the number of computations in a layer.

At a low level, many neural network layers can be viewed as repeated weighted sums. A fully connected neuron computes a dot product between an input vector and a weight vector, while a convolutional output can be written as a dot product between an unfolded input patch and a flattened kernel. In standard networks, the resulting pre-activation is passed through a nonlinear activation function and may take a wide range of values. Binarizing the activations changes this setting. Restricting the output activation to two possible values makes only the sign of the pre-activation relevant, rather than its exact magnitude. For a binary input activation $a_i\in\{-1,+1\}$, each product $a_iw_i$ becomes a sign-controlled accumulation that either adds or subtracts the real-valued weight $w_i$. Binarization therefore reduces the cost of each term, but still evaluates every term of the sum. 

This raises a simple question: \textit{are all $N$ terms always necessary to determine the final binary output?}

For some inputs, the accumulated value may become strongly positive or strongly negative before all terms have been evaluated. In such cases, the trajectory taken by the sum may already provide sufficient evidence of the final sign. The remaining terms can then be skipped, reducing computation within the accumulation itself. This paper studies this possibility through a post-training early-stopping mechanism for binary-activation layers. The method monitors partial sums during accumulation and, at selected checkpoints, decides whether computation can stop without evaluating the remaining terms.  We study threshold calibration, checkpoint placement, and the resulting accuracy--computation trade-off. The scope is deliberately algorithmic and hardware-agnostic. Savings are measured as evaluated input--weight contributions relative to dense accumulation, under an ideal execution model in which output units may follow independent accumulation orders and stop independently. We therefore report arithmetic savings rather than latency, memory traffic, or energy. Our principal contributions are:
\begin{itemize}
    \item A post-training mechanism that determines binary outputs from ordered partial sums before completing fully connected or convolutional accumulations.
    
    \item Empirical-quantile and parametric-Gaussian threshold calibration, with configurable threshold granularity and fixed or data-driven checkpoint schedules.
    
    \item An evaluation on binary-activation VGG11 and CIFAR-10, covering convolutional and fully connected layers and reporting  local and full-network arithmetic savings.
\end{itemize}

The remainder of the paper is organized as follows. 
Section~\ref{sec:related} reviews network binarization, dynamic inference, and pruning, while Sections~\ref{sec:background} and~\ref{sec:method} formalize the early-stopping problem and present threshold calibration and checkpoint scheduling. Section~\ref{sec:setup} describes the experimental protocol, followed by the results and discussion.
\section{Related Work}\label{sec:related}

\subsection{Quantized and Binary Networks}
BinaryConnect~\cite{courbariaux2015binaryconnect} introduced an extreme form of quantization by using binary weights during training and inference to reduce model storage and simplify arithmetic. Binarized Neural Networks~\cite{courbariaux2016bnn} extended this principle to both weights and activations, while XNOR-Net~\cite{rastegari2016xnor} further adapts this principle to convolutional networks. Binary filters and binary inputs together with scaling factors allow convolutional operations to be approximated with efficient bitwise operations.

These approaches mainly reduce the storage cost and complexity of each elementary operation, but every input contribution of a layer is still evaluated.

\subsection{Dynamic Inference and Partial Computation}
Dynamic inference methods adapt the executed computation to each input. BranchyNet~\cite{teerapittayanon2016branchynet} adds intermediate exit branches to a network: when an early classifier is sufficiently confident, for example when its output entropy is low, inference can stop before reaching the final layer. SkipNet~\cite{wang2018skipnet} instead learns to skip architectural components such as residual blocks, allowing easier inputs to traverse fewer layers. At a finer granularity, Feature Boosting and Suppression (FBS)~\cite{gao2019dynamicchannel} uses a lightweight predictor to estimate which feature channels are salient for the current input and dynamically suppress less useful ones.

Static pruning instead removes parameters or channels permanently, from Optimal Brain Damage~\cite{lecun1989optimal} to Deep Compression~\cite{han2016deepcompression}; the resulting structure is therefore shared by all inputs. Closer to our setting, Incomplete Dot Products (IDP)~\cite{mcdanel2017idp} reduce computation \emph{inside} a layer by evaluating only a prefix of the input channels. However, this prefix follows a preset computation budget rather than each output's partial-sum trajectory, and the network must be trained to concentrate information in its early channels. Our method instead makes input-dependent stopping decisions after training.

\subsection{Early Termination of Accumulations}
The closest prior work uses the fact that ReLU maps negative pre-activations to zero to terminate an accumulation once its outcome becomes predictable. SnaPEA~\cite{akhlaghi2018snapea} reorders weights by sign to expose negative outputs earlier: its exact mode stops when the remaining contributions cannot restore a positive result, whereas its predictive mode compares the partial sum with an offline-calibrated threshold. CompRRAE~\cite{chen2019comprrae} estimates the contribution of remaining bit planes, while ConvReLU++~\cite{kong2023convrelu} derives reference-based bounds to identify dot products whose ReLU output is guaranteed to be zero.

We address the same early-termination problem for layers with binary input and output activations, where the sign of the final pre-activation determines the output. Rather than detecting only negative ReLU outputs, our rule can predict either binary sign.
\section{Background and Problem Formulation}\label{sec:background}

\subsection{Binary-Output Trajectories and Reordering}

Consider a single binary-output neuron with $N$ binary input activations $a_i\in\{-1,+1\}$, real-valued weights $w_i$, and a bias $b$; the neuron computes a weighted sum and returns only its sign. After $k$ terms accumulated, the partial sum is $S_k = b + \sum_{i=1}^{k} a_iw_i$ ($0\leq k\leq N$), and the binary output depends only on the sign of the final value $S_N$. The sequence $(S_0,S_1,\ldots,S_N)$ forms the partial-sum trajectory of the neuron for a particular input.

\begin{figure*}[t]
    \centering
    \includegraphics[width=\linewidth]{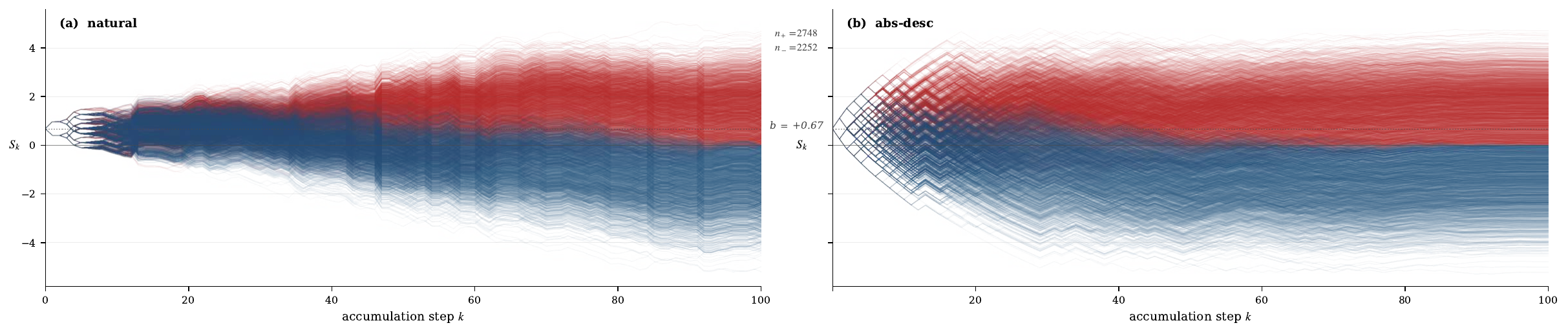} 
    \caption{Partial-sum trajectories for one neuron ($u = 50$) in the first hidden layer of a binary-activation MLP trained on MNIST, obtained from 5000 input samples with $N=100$. Red trajectories are $\operatorname{sign}(S_N)=+1$, blue trajectories satisfy $\operatorname{sign}(S_N)=-1$. \textbf{(a)} Natural ordering. \textbf{(b)} Descending-magnitude (\textit{abs-desc}) ordering.}
    \label{fig:mnist-mlp-trajectories}
\end{figure*}

Figure~\ref{fig:mnist-mlp-trajectories}(a) shows trajectories of 5000 samples. Each curve corresponds to a different input sample and is colored according to the sign of its complete accumulation. Trajectories with positive and negative terminal signs remain strongly overlapped during most of the accumulation, making an early prediction of the final sign difficult. This observation raises a question: \textit{can we make the sign visible earlier?}

Let $\pi$ denote a permutation of the $N$ input terms where the natural order corresponds to $\pi(j)=j$. 
A more appropriate order is to evaluate the weights in descending magnitude:
\begin{equation}
    |w_{\pi(1)}|
    \geq |w_{\pi(2)}|
    \geq \cdots
    \geq |w_{\pi(N)}|.
    \label{eq:abs-ordering}
\end{equation}
Since $a_i\in\{-1,+1\}$, each contribution satisfies $|a_iw_i|=|w_i|$. Its magnitude is therefore known from the weight alone, independently of the current input.

Figure~\ref{fig:mnist-mlp-trajectories}(b) shows the trajectories obtained with this descending-magnitude order. Compared with the natural order in Fig.~\ref{fig:mnist-mlp-trajectories}(a), the positive and negative groups separate earlier because the largest-magnitude contributions are accumulated first. The terminal sign may therefore become predictable before all $N$ terms have been evaluated.

Reordering only reshapes the trajectory from which an early decision may be made. Computation is saved only if a stopping strategy can determine, from the available partial information, that the remaining terms are unnecessary. The following subsection generalizes this trajectory and ordering from a single MLP neuron to fully connected and convolutional binary-output operators.

\subsection{Unified Binary-Output Operator}
We consider layers with binary input activations and binary output activations. For a fully connected layer, an observation $r$ is a single input vector, and an output unit $u$ is a single neuron. In a convolution, an observation $r$ is an unfolded input patch at a given spatial location, and an output unit $u$ is an output channel.

In both cases, each output unit computes a dot product over $N$ input terms where $N$ is the input dimension for a fully connected layer, whereas $N=C_{\mathrm{in}}k_hk_w$ for a 2-D convolution, where $C_{\mathrm{in}}$ is the number of input channels and $k_h\times k_w$ are the kernel height and width, respectively.

As defined earlier, $\pi_u$ is the evaluation order of the $N$ input terms for output unit $u$. The accumulation after $k$ terms is
\begin{equation}
    S_{u,k}^{(r)} = b_u + \sum_{j=1}^{k} a_{\pi_u(j)}^{(r)} w_{u,\pi_u(j)},
    \begin{aligned}
        \text{with }0\leq k\leq N, \\
        \quad a_i^{(r)}\in\{-1,+1\}.
    \end{aligned}
    \label{eq:partial-sum}
\end{equation}
Here $S_{u,0}^{(r)}=b_u$ and $S_{u,N}^{(r)}$ is the complete accumulation, whose sign gives the binary output $y_u^{(r)}=\operatorname{sign}(S_{u,N}^{(r)})$ ($\operatorname{sign}(z)=+1$ for $z\geq0$ and $-1$ otherwise).

\subsection{Exact Remaining Bound and Opportunity}
\label{sec:exact-bound}

With binary inputs, the magnitude of the unevaluated terms is bounded independently of the input:
\begin{equation}
    R_{u,k}
    =
    \sum_{j=k+1}^{N}|w_{u,\pi_u(j)}|,
    \qquad
    \left|S_{u,N}^{(r)}-S_{u,k}^{(r)}\right|
    \leq R_{u,k}.
    \label{eq:exact-remaining}
\end{equation}
Therefore, if $S_{u,k}^{(r)}>R_{u,k}$, the final sign must be positive; if $S_{u,k}^{(r)}<-R_{u,k}$, it must be negative, since the remaining terms cannot move the complete sum across zero. This exact rule is deterministic but conservative: the partial-sum sign may already match the final sign while $|S_{u,k}^{(r)}|\leq R_{u,k}$.

We exploit this opportunity through two complementary components: an ordering $\pi_u$, which reshapes the partial-sum trajectories without reducing computation by itself, and a stopping policy, which saves computation whenever a decision is reached before step $N$. Descending-magnitude ordering is used as the default preprocessing choice. The central problem is therefore to determine which decision rules to apply to the partial sums and when to apply them.

\section{Threshold-Based Early Stopping}\label{sec:method}

\subsection{Assumptions}

The analysis targets operators with binary inputs, real-valued effective weights, and binary outputs obtained by applying a sign function to the complete pre-activation. Biases and inference-time batch-normalization parameters are folded into the affine operator. The first layer of a network is therefore excluded when its inputs are not binary, and the present formulation is not applied to a final real-valued classifier.

\subsection{Threshold-Based Dynamic Rule}
\label{sec:decision-rule}

The exact remaining-bound rule in~(\ref{eq:exact-remaining}) provides a deterministic conservative stopping condition because it accounts for the worst possible contribution of all remaining terms. At a given step $k$, a large positive partial sum should indicate that the complete accumulation will end with a positive sign, while a large negative partial sum should indicate a negative sign. Values that are not clearly associated with either sign remain inside an uncertainty band. We therefore introduce an empirical threshold-based dynamic rule that can terminate an accumulation when its partial sum provides sufficient evidence for either binary outcome.

Let $\mathcal{K}\subseteq\{1,\ldots,N-1\}$ denote the set of selected checkpoint steps. At a checkpoint $k\in\mathcal{K}$, the decision $d_{u,k}^{(r)}$ is defined as
\begin{equation}
    d_{u,k}^{(r)} =
    \begin{cases}
        +1, & S_{u,k}^{(r)}
               > T_\mathrm{high}^{(k)},\\
        -1, & S_{u,k}^{(r)}
               < T_\mathrm{low}^{(k)},\\
        0,  & \text{otherwise}.
    \end{cases}
    \label{eq:decision-rule}
\end{equation}
The lower $T_\mathrm{low}^{(k)}$ and upper $T_\mathrm{high}^{(k)}$ thresholds define an uncertainty band at every step $k$. A partial sum outside this band produces an early binary decision, whereas $d_{u,k}^{(r)}=0$ indicates that the accumulation must continue.

Checkpoints are evaluated in increasing order, and the first nonzero decision fixes the output and skips all remaining terms. If no early decision is reached, the complete accumulation is evaluated and the output is determined by $\operatorname{sign}(S_{u,N}^{(r)})$.
Figure~\ref{fig:mnist-checkpoints} illustrates the complete mechanism on one output unit: at each checkpoint, the trajectories outside the calibrated band stop with an early decision, while the sign-conditioned populations that remain inside continue accumulating.

\subsection{Threshold Calibration}\label{sec:calibration}

Threshold calibration relies on the partial sums computed on the training dataset and their corresponding final signs to construct empirical decision boundaries. Calibration uses a fixed subset $\mathcal{D}_{\mathrm{cal}}$ of the training partition (Section~\ref{sec:setup}); the samples need not be held out from model training, since no parameter is updated. On $\mathcal{D}_{\mathrm{cal}}$ the dense model collects $S_{u,k}^{(r)}$ and its reference sign $y_u^{(r)}$ at every step. For clarity, calibration is described here for a single output unit $u$; sharing thresholds across several units is deferred to Section~\ref{sec:granularity}. For each output unit $u$ and step $k$, the calibration trajectories are split in $2$ according to the dense reference sign $\pm$:
\begin{equation}
    \mathcal{S}_{u,k}^{\pm}
    = \left\{
      S_{u,k}^{(r)}:
      y_u^{(r)}=\pm1
      \right\}
    \label{eq:conditional-samples}
\end{equation}
For a convolution, $r$ denotes an image-position pair, so calibration pools all spatial locations across images: $100$ images with an $8\times8$ feature map provide $6{,}400$ observations per output channel. The class label is not used; conditioning is based only on the dense binary output. 

Each calibration method produces two candidate cut points: $c_{u,k}^{-}$ from the negative population and $c_{u,k}^{+}$ from the positive population. These candidates derive the lower and upper threshold,
\begin{align}
    T_{\mathrm{low},u}^{(k)}=\min(c_{u,k}^{-},c_{u,k}^{+}), ~~T_{\mathrm{high},u}^{(k)}=\max(c_{u,k}^{-},c_{u,k}^{+})
    \label{eq:sorted-thresholds}
\end{align}
This construction ensures that a valid uncertainty interval $[T_{\mathrm{low},u}^{(k)},T_{\mathrm{high},u}^{(k)}]$ is always obtained. A partial sum above is assigned to the positive sign, a partial sum below is assigned to the negative sign, and values inside continue accumulating.

To clarify the role of $\min/\max$, consider the case in which $c_{u,k}^{-}$ is the largest partial sum observed among negative examples and $c_{u,k}^{+}$ the smallest among positive examples. If $c_{u,k}^{+}<c_{u,k}^{-}$, the two populations overlap; if $c_{u,k}^{+}>c_{u,k}^{-}$, they are separated and the interval between the candidates forms an unobserved gap. In either case, the candidate cut points yield a valid uncertainty band, illustrated in Fig.~\ref{fig:mnist-checkpoints}(a), within which accumulation continues. We next consider two ways to compute these candidate cut points.

\subsubsection{Empirical Quantile Calibration}

Let $Q_q(\mathcal{S})$ denote the empirical quantile of sample $\mathcal{S}$ at level $q$. For quantile levels $(q_+,q_-)$, we compute one candidate from each reference-sign population:
\begin{align}
    c_{u,k}^{-} &= Q_{q_-}\!\left(\mathcal{S}_{u,k}^{-}\right),&
    c_{u,k}^{+} &= Q_{q_+}\!\left(\mathcal{S}_{u,k}^{+}\right)
    \label{eq:quantile-candidates}
\end{align}
In our experiments, we use a symmetric configuration with $q_+=\alpha$ and $q_-=1-\alpha$; for example $\alpha=0.05$ gives $(q_+, q_-) = (0.05,0.95)$. On the calibration observations, values from the negative-reference population above $c_{u,k}^{-}$ occupy approximately its upper $\alpha$ tail, while positive-reference values below $c_{u,k}^{+}$ occupy approximately its lower $\alpha$ tail.

\subsubsection{Parametric Gaussian Calibration}

Each sign-conditioned population is approximated by a Gaussian with mean $\mu_{u,k}^{s}$ and standard deviation $\sigma_{u,k}^{s}$. Using the standard Gaussian quantile $z_{\ptarget}=\Phi^{-1}(1-\ptarget)$, the two candidates are:
\begin{align}
    c_{u,k}^{-}
        &=\mu_{u,k}^{-}+z_{\ptarget}\sigma_{u,k}^{-},\\
    c_{u,k}^{+}
        &=\mu_{u,k}^{+}-z_{\ptarget}\sigma_{u,k}^{+}
    \label{eq:gaussian-candidates}
\end{align}
The parameter $\ptarget$ controls how much probability mass is left outside the candidate threshold in each fitted tail. For example, $\ptarget=0.05$ places the negative candidate $c_{u,k}^{-}$ at the $95$th percentile of the fitted negative distribution and the positive candidate $c_{u,k}^{+}$ at the $5$th percentile. 

The final thresholds are again obtained from \eqref{eq:sorted-thresholds}. Under the fitted Gaussian model, each candidate cuts a tail of mass $\ptarget$ from its corresponding conditional distribution. This interpretation depends on the adequacy of the Gaussian approximation and is not a formal error guarantee.

\begin{figure*}[t]
    \centering
    \includegraphics[width=\linewidth]{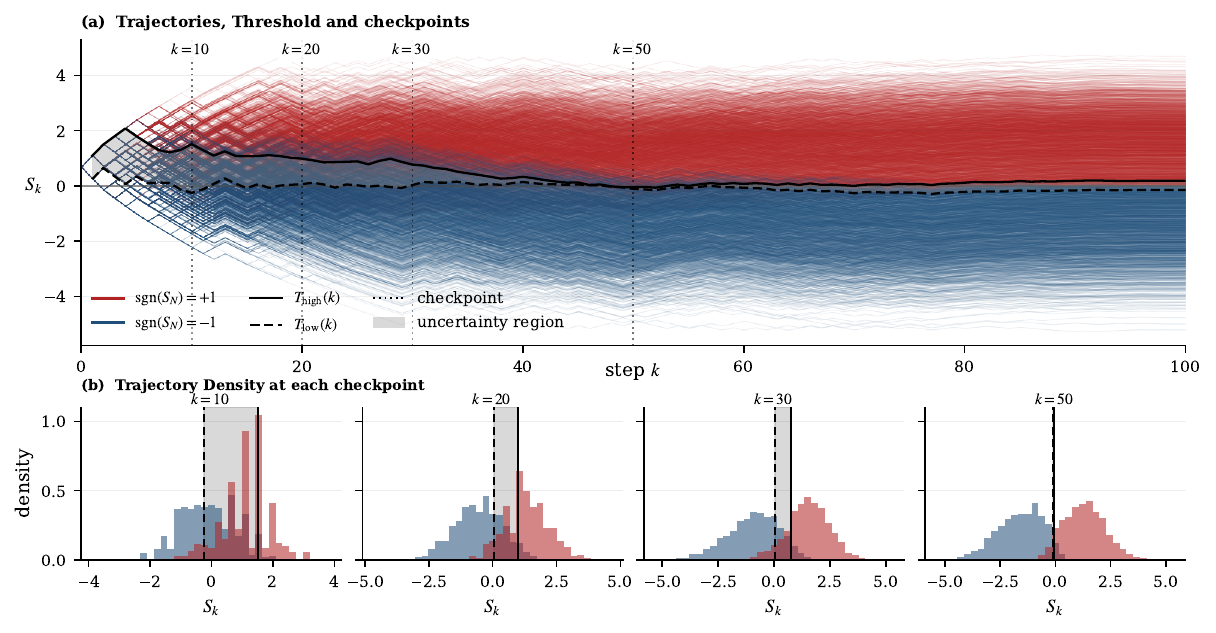}
    \caption{Threshold-based early stopping illustrated on the same setup as Fig. \ref{fig:mnist-mlp-trajectories} ($u = 50$, $N=100$, $5{,}000$ inputs). \textbf{(a)} Partial-sum trajectories $S_k$ under abs-desc ordering, coloured by their final sign ($\operatorname{sgn}(S_N)=+1$ in red, $-1$ in blue). The black curves are the per-unit thresholds $\Thigh(k)$ (solid) and $\Tlow(k)$ (dashed), calibrated on the calibration split with empirical quantiles ($\alpha=0.05$). Dotted verticals mark the checkpoints of the \texttt{percent\_4} schedule ($k\in\{10,20,30,50\}$). \textbf{(b)} Distribution of $S_k$ over all test inputs at each checkpoint, with the local threshold band.}
    \label{fig:mnist-checkpoints}
\end{figure*}

Since thresholds are tested only at selected checkpoints, the method must still return a binary output when no checkpoint fires. We therefore define a static fallback step $k_u^\star$, the latest step at which the accumulation may stop. The conservative $k_u^\star=N$ completes the accumulation and recovers the dense sign (up to floating-point reordering), whereas a more aggressive $k_u^\star<N$ decides directly from the partial sum at that step, with $\hat{y}_{u}^{(r)}=\operatorname{sign}\!\bigl(S_{u,k_u^\star}^{(r)}-c_{u}^{(k_u^\star)}\bigr)$, using $c_u^{(k)}$ as center of the threshold band.

A good fallback position is a step where the positive and negative reference populations are separated, i.e.\ where the signed margin $c_{u,k}^{+}-c_{u,k}^{-}$ is positive; since this margin is only an empirical diagnostic computed on calibration data, any $k_u^\star<N$ is treated as a hyperparameter evaluated through the accuracy--computation trade-off.

\subsection{Checkpoint Schedules}\label{sec:checkpoint}

A checkpoint schedule $\mathcal{K}_{\ell,u}$ specifies when the partial sum is compared with the calibrated thresholds before the fallback step. Checkpoints are evaluated in increasing order. Adding checkpoints creates more opportunities to stop, but also increases comparison overhead and risk of early decisions.

\paragraph{Data-independent schedules.}
These schedules do not use calibration trajectories. A \textbf{stride} schedule tests every $s$-th step, whereas a \textbf{fixed or percentage} schedule tests user-selected positions such as $\lceil pN_\ell\rceil$. For example, \texttt{percent\_4} tests at $\{10,20,30,50\}\%$ of $N$ (Fig.~\ref{fig:mnist-checkpoints}).

\paragraph{Data-driven schedules.}\label{subsec:data-driven_policices}
Data-driven schedules are constructed from a reference stop step $\tau$ computed for each calibration observation and used only to select checkpoint positions. We consider the \emph{calibrated crossing}, the first step at which a threshold would have fired; the \emph{exact remaining-bound}, the first step satisfying \eqref{eq:exact-remaining}; and the \emph{sign-lock}, the first step after which the partial-sum sign remains equal to the final sign. A \textbf{cumulative distribution function (CDF)} schedule reads its checkpoints from the empirical distribution of reference stop steps, that is, the fraction of observations whose stop step is at most $k$. It places checkpoints at the smallest steps where it reaches the chosen target levels; for example \texttt{cdf\_60\_90} uses the earliest steps by which $60\%$ and $90\%$ of the reference stops have occurred. A \textbf{moment} schedule instead uses the mean and std of the same stop-step distribution, placing checkpoints at $\mu_\tau+m\sigma_\tau$.

\subsection{Calibration Granularity}\label{sec:granularity}

Thresholds may be calibrated per output unit or shared by a complete layer. Per-unit calibration gives each neuron or convolutional output channel its own decision band; for convolutions it is shared across the spatial positions of that channel. Layer-level calibration instead pools all conditional observations in the layer, reducing storage of thresholds at the cost of mixing units with different weight scales. Threshold granularity (\emph{which} band) is configured independently of checkpoint granularity (\emph{when} it is tested). Our experiments use per-unit thresholds with layer-shared checkpoints, retaining unit-specific calibration while limiting control overhead.

\section{Experimental Setup}\label{sec:setup}
\label{sec:experiments}

We evaluate the method on a binary-activation VGG11 adapted to CIFAR-10~\cite{krizhevsky2009cifar}, with binary convolutional blocks and a single binary fully connected block ($2048\!\to\!1024$) replacing the original multi-layer classifier, allowing both operator types to be studied under the same protocol. Only layers with binary inputs are thresholded: the first image-input layer and the final real-valued classifier remain unchanged, while batch-normalization parameters are folded into each supported operator. All experiments are post-training, with frozen weights and no fine-tuning. For each target layer, ordered partial sums and dense reference signs are collected to calibrate thresholds and checkpoints.

The model is trained on 45k images, of which 5k are reused for calibration; the remaining 5k CIFAR-10 training images are reserved for policy selection, and the official 10k test set is used only for final reporting. Results are averaged over three independently trained seeds ($42$, $43$, and $44$). In multi-layer experiments, selected layers are thresholded jointly, allowing upstream early decisions to propagate downstream. All experiments use PyTorch and run on NVIDIA L40S GPU.

\paragraph{Policies and metrics.}
Accumulation terms are ordered per output unit by descending absolute weight magnitude~\eqref{eq:abs-ordering}, with full accumulation as fallback ($k_u^\star=N$). We use per-unit thresholds, layer-shared checkpoints, both calibration modes of Section~\ref{sec:calibration} and checkpoint schedules of Section~\ref{sec:method}.

The primary cost is the average number of evaluated input--weight contributions, counted as hardware-independent MAC-equivalents. These counts assume that output units can follow independent accumulation orders and stop independently; they therefore represent ideal arithmetic savings rather than measured latency, memory traffic, or energy. 

Let $\overline{C_{\mathcal P}^{\mathcal L}}$ denote the average number of terms evaluated by policy $\mathcal P$ in the targeted layers $\mathcal L$. We report
\begin{equation}
\begin{aligned}
r_{\mathrm{local}}
&=
1-
\frac{\overline{C_{\mathcal P}^{\mathcal L}}}
     {C_{\mathrm{dense}}^{\mathcal L}},
&
r_{\mathrm{arch}}
&=
\frac{C_{\mathrm{dense}}^{\mathcal L}}
     {C_{\mathrm{dense}}^{\mathrm{full}}}
r_{\mathrm{local}} .
\end{aligned}
\label{eq:reductions}
\end{equation}
The first measures savings inside the targeted layers, whereas the second rescales it to the full-network budget.

Accuracy is measured against the original dense model and, when relevant, against the reordered dense reference to isolate floating-point reordering drift~\cite{goldberg1991floating} from threshold-induced errors. The number of checkpoint tests is also reported as a control-cost diagnostic, without assigning it a hardware cost.
\section{Results}\label{sec:results}

All numbers are single test-set evaluations reported after policy selection on the validation split. Accuracy drops are in percentage points (pp) against the dense baseline, and the reduction ratios $r_{\mathrm{local}}$ and $r_{\mathrm{arch}}$ follow \eqref{eq:reductions}.

\subsection{Single-Layer Thresholding}\label{sec:res-single}

The first experiment thresholds only the deepest convolution of VGG11, \texttt{features.7} ($512$ output channels, $N=4{,}608$ terms per output, about $3.2\%$ of the dense budget); its results form the F7 block of Table~\ref{tab:main-results}. The high-reduction schedule \texttt{percent\_4} skips $86.58 \pm 0.04\%$ of the layer's contributions and keeps a test accuracy of $88.08 \pm 0.15\%$, a $0.37 \pm 0.06$\,pp drop from the $88.45\%$ dense baseline; the accuracy cost rises smoothly with aggressiveness: \texttt{mean\_std\_2} gives up three points of local reduction for a slightly smaller drop, while the \texttt{cdf} variants ($+0.46$\,pp at $80.6\%$ and $+0.87$\,pp at $86.3\%$) fill the front between them. Calibrating the same layer with the parametric Gaussian mode instead of empirical quantiles is more conservative at matched risk: its three-seed points skip $70$--$75\%$ of the contributions for drops of only $0.02$--$0.12$\,pp (\texttt{mean\_std\_2} and \texttt{percent\_4}), reaching near-lossless operation from just the per-population means and variances.

\begin{table*}[t]
\caption{Complete test-set thresholding results (empirical calibration,
$\alpha=0.05$; mean $\pm$ across-seed std, three seeds).
\emph{Full} thresholds all eight analyzable Conv and FC blocks at once.
Acc.\ is the measured thresholded test accuracy and Drop is the dense
baseline $A_{\mathrm{orig}}=88.45\%$ ($\pm0.14$) minus Acc.;
the across-seed standard deviation of $r_{\mathrm{local}}$ is $\leq 0.14$.}
\label{tab:main-results}

\centering
\small
\setlength{\tabcolsep}{6pt}

\begin{tabular*}{\textwidth}{
    @{\extracolsep{\fill}}
    l l r r r r
    @{}
}
\toprule
\textbf{Layer set} &
\textbf{Policy} &
\textbf{Acc.\ (\%)} &
\textbf{Drop (pp)} &
\textbf{$r_{\mathrm{local}}$ (\%)} &
\textbf{$r_{\mathrm{arch}}$ (\%)} \\
\midrule

F7 & \texttt{mean\_std\_2}
& \textbf{88.13} & $\mathbf{+0.32 \pm 0.12}$ & 83.38 & 2.65 \\

F7 & \texttt{percent\_4}
& 88.08 & $+0.37 \pm 0.06$ & \textbf{86.58} & \textbf{2.76} \\

\midrule

FC & \texttt{percent\_4}
& \textbf{88.04} & $\mathbf{+0.41 \pm 0.17}$ & 89.26 & 0.63 \\

FC & \texttt{cdf\_40\_75\_90}
& 87.94 & $+0.51 \pm 0.09$ & \textbf{92.21} & \textbf{0.65} \\

\midrule

F7+FC & \texttt{mean\_std\_2}
& \textbf{87.65} & $\mathbf{+0.76 \pm 0.04}$ & 84.79 & 3.30 \\

F7+FC & \texttt{percent\_4}
& 87.53 & $+0.92 \pm 0.09$ & 87.06 & 3.39 \\

F7+FC & \texttt{cdf\_40\_75\_90}
& 86.77 & $+1.68 \pm 0.34$ & \textbf{87.41} & \textbf{3.40} \\

\midrule

F5+F6+F7 & \texttt{mean\_std\_2}
& \textbf{87.16} & $\mathbf{+1.25 \pm 0.13}$ & 84.69 & 24.27 \\

F5+F6+F7 & \texttt{percent\_4}
& 87.09 & $+1.36 \pm 0.22$ & 87.08 & 24.95 \\

F5+F6+F7 & \texttt{cdf\_40\_75\_90}
& 85.15 & $+3.30 \pm 0.30$ & \textbf{87.93} & \textbf{25.19} \\

\midrule

Full & \texttt{mean\_std\_2}
& \textbf{82.36} & $\mathbf{+6.09 \pm 0.54}$ & 83.62 & 83.12 \\

Full & \texttt{percent\_4}
& 81.04 & $+7.41 \pm 0.54$ & 86.55 & 86.03 \\

Full & \texttt{cdf\_60\_90}
& 79.09 & $+9.36 \pm 0.54$ & 80.90 & 80.42 \\

Full & \texttt{cdf\_40\_75\_90}
& 73.61 & $+14.84 \pm 0.86$ & \textbf{86.72} & \textbf{86.20} \\

\bottomrule
\end{tabular*}
\end{table*}

The stop-step references of Section~\ref{sec:checkpoint} clarify this result. On \texttt{features.7}, the median \emph{realised sign-lock} occurs at step $77$ ($1.7\%$ of $N$), while the deterministic \emph{exact bound} fires only at step $3{,}789$ ($82.2\%$). The calibrated \texttt{percent\_4} policy exploits this gap, skipping $86.6\%$ of contributions versus about $18\%$ under the exact rule.

\subsection{Fully Connected and Multi-Layer Sets}\label{sec:res-mixed}

The remaining blocks of Table~\ref{tab:main-results} add three configurations of growing coverage: the binary FC block alone, the joint F7$+$FC pair, and the three deepest convolutions F5$+$F6$+$F7.

On the FC block alone, the local reduction reaches $89$--$92\%$ for drops below $0.6$\,pp, but its dense budget is small (about $2$\,M operations), so the architecture-level effect stays modest ($r_{\mathrm{arch}} \approx 0.65\%$). Adding the FC block on top of \texttt{features.7} costs a further $0.5$--$0.8$\,pp.

The multilayer set carries the architecture-level claim: three contiguous deep convolutions lift $r_{\mathrm{arch}}$ to about $25\%$ of the full-network arithmetic, with \texttt{percent\_4} reaching $87.09 \pm 0.09\%$ test accuracy for $24.95\%$ of the budget, a $1.36 \pm 0.22$\,pp drop from the dense baseline.

We also report the number of checkpoint tests as a proxy for control cost, since thresholds are tested at a few checkpoints rather than after every term. On F5+F6+F7, \texttt{percent\_4} makes $2.35\times10^{4}$ threshold comparisons per input ($0.03\%$ of the dense term count) to skip $7.4\times10^{7}$ terms, i.e.\ about $3{,}100$ terms removed per comparison.

\subsection{Trading Accuracy for Larger Savings}\label{sec:res-sweep}

\begin{figure}[t]
    \centering
    \includegraphics[width=.97\linewidth]{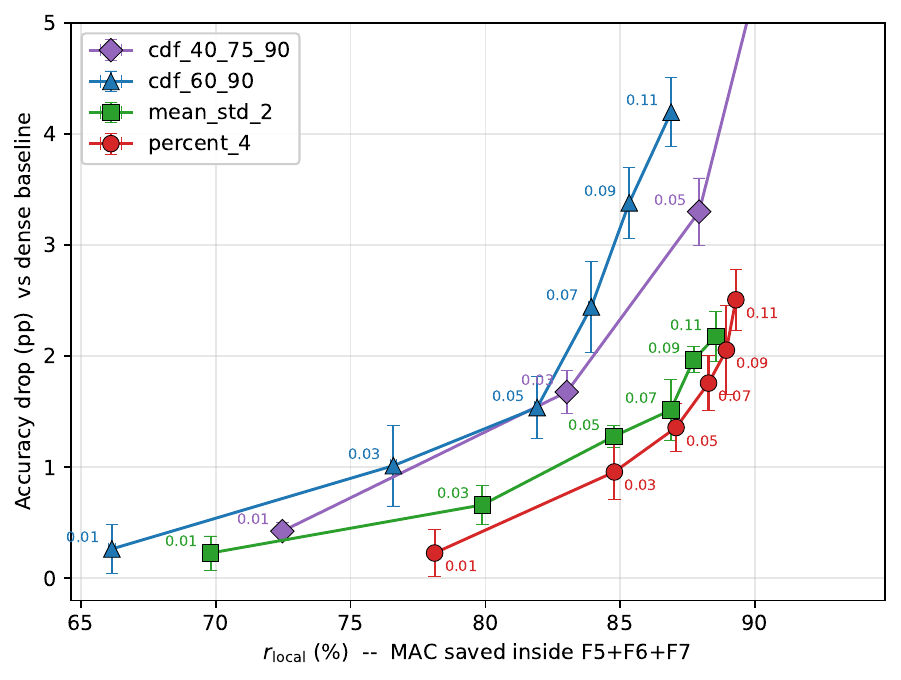}
    \caption{Accuracy--computation on F5+F6+F7, sweeping the empirical calibration quantile $\alpha \in \{0.01,\dots,0.11\}$ (annotated next to each marker) for the four schedule families.}
    \label{fig:multilayer-sweep}
\end{figure}

To further show how far the mechanism can be pushed, we sweep the empirical calibration quantile $\alpha$ from $0.01$ to $0.11$ on F5+F6+F7 and trace the resulting accuracy--computation (Figure~\ref{fig:multilayer-sweep}). A conservative threshold ($\alpha=0.01$) already skips $78\%$ of the three layers, about $22\%$ of the whole-network computation, for a $0.23$\,pp drop, and the two best families (\texttt{percent\_4} and \texttt{mean\_std\_2}) stay below $1.5$\,pp up to $\alpha=0.05$ ($87\%$ local). The CDF families are steeper: past $\alpha=0.05$ their drop grows quickly for little extra reduction.

The second experiment thresholds \emph{every} binary layer of the network, that is, all eight analyzable Conv and FC blocks at once, and so measures the ceiling of the method (the \emph{Full} block of Table~\ref{tab:main-results}). At $\alpha=0.05$ the whole-network arithmetic can be cut by $83$--$86\%$, but the cascade of eight approximate layers now costs $6$--$15$\,pp of accuracy. The moment schedule \texttt{mean\_std\_2} is the best point, reaching $82.36 \pm 0.47\%$ test accuracy, $6.09$\,pp below the dense baseline, while removing $83.1\%$ of the network's arithmetic.

\subsection{Per-Channel Accumulation Ordering}\label{sec:res-ordering}

The headline results use per-channel \emph{abs-desc} ordering, which evaluates each channel's largest-magnitude contributions first. To measure how much the savings rely on it, we re-run the single-layer F7 analysis (seed $42$, $\alpha=0.05$) under \emph{natural} ordering, recalibrating the thresholds for each order, and Figure~\ref{fig:mnist-mlp-trajectories} shows why natural ordering is harder, keeping the sign-conditioned populations overlapped far longer.

Switching to natural ordering costs $22$--$36$\,pp of local reduction across policies. The fixed \texttt{percent\_4}, whose early checkpoints are tuned to the abs-desc concentration, is hit hardest: its accuracy drop grows from $0.43$ to $4.56$\,pp while its reduction falls from $86.6\%$ to $64.7\%$. The data-driven schedules degrade far more gracefully, staying below $1.6$\,pp at $44$--$51\%$ reduction. The effect compounds in cascade: on F5+F6+F7 under natural ordering, every schedule drops by more than $5.5$\,pp, \texttt{percent\_4} by $24.8$\,pp. This sets the price of the per-channel abs-desc assumption (Section~\ref{sec:disc-limits}).

\section{Discussion}\label{sec:discussion}

\subsection{Interpreting the Reported Savings}\label{sec:disc-reorder}

The reported accuracy losses arise from the threshold decisions, not from the per-channel reordering. Across every layer set the reordered dense reference $A_{\mathrm{reord}}$ departs from the original $A_{\mathrm{orig}}$ by at most $0.09$\,pp ($0.00$ on F7/FC/F7+FC, $-0.03\pm0.06$ on F5+F6+F7, $+0.09\pm0.08$ on Full).

Savings must nevertheless be interpreted at two scales. The \texttt{percent\_4} policy removes $86.6\%$ of F7's contributions, but only $2.76\%$ of the full-network arithmetic because F7 represents about $3.2\%$ of the dense budget. Applying the mechanism to F5+F6+F7 preserves a similar local reduction while increasing the architecture-level saving to $24.95\%$. Local reductions should therefore be reported alongside the computational share of the targeted layers, since $r_{\mathrm{local}}$ alone can overstate the impact of the method.

\subsection{Choosing a Calibration and a Schedule}\label{sec:disc-calib}

The choice of calibration mainly depends on the amount of available data. Empirical quantiles estimate the decision band directly from the tails of the sign-conditioned partial-sum distributions and therefore benefit from many observations per threshold group. The Gaussian model summarizes each population using only a mean and variance, reducing calibration storage, but produces more conservative operating points in our experiments. Empirical calibration is thus preferable when calibration data are plentiful, whereas parametric provides a lighter alternative when data are limited.

Checkpoint placement also depends strongly on the accumulation order. Under abs-desc ordering, the largest-magnitude terms are evaluated first, causing the final sign to emerge early; a few fixed checkpoints such as those of \texttt{percent\_4} can therefore capture most decisions. Under natural ordering, decisive contributions are more dispersed and sign separation occurs gradually, making data-driven \texttt{cdf} schedules more effective than the same fixed positions. Fixed schedules are convenient when the stop-step distribution matches their assumed pattern, but checkpoint positions should otherwise be derived from that distribution for the selected accumulation order and computation budget.

\subsection{Limitations and Outlook}\label{sec:disc-limits}

The main limitation is the unit-specific abs-desc ordering, which is not directly compatible with a single shared-input fetch. The reported reductions should therefore be interpreted as idealized arithmetic savings rather than predictions of achievable hardware latency. We count MAC-equivalent contributions but do not model memory traffic, synchronization, branching, or the energy and latency cost of checkpoint tests. A deployable implementation will require shared or grouped accumulation orders, a hardware-aware cost model, and potentially co-design with a binary accelerator.

The evaluation is also limited to a moderate-scale model. Although the thresholding mechanism itself is post-training and independent of a particular architecture, extending it to deeper binary ResNets or vision and language transformers requires suitable pretrained binary models and calibration procedures that scale beyond per-unit trajectory buffers.

Finally, thresholding several layers introduces cascading errors: an early decision changes the input received by subsequent thresholded layers, allowing a local sign error to propagate. The reported accuracies already include this effect because policies are evaluated through complete thresholded forward passes. However, each layer is calibrated independently from the dense reference, and the model is not fine-tuned for thresholded inference. Adapting thresholds to upstream decisions or lightly fine-tuning the network may therefore recover part of the lost accuracy. Despite these limitations, the policy ranking remains consistent across the three training seeds, with an across-seed standard deviation of the accuracy drop below $0.34$\,pp. With only three seeds, these results indicate stability rather than statistical confidence.

\section{Conclusion}\label{sec:conclusion}

We presented a post-training mechanism that stops a binary-activation accumulation once its partial sum leaves a calibrated decision band, predicting the final sign without completing the sum or retraining the model. Applying it to fully connected and convolutional operators on a binary-activation VGG11 for CIFAR-10 removes $86.6\%$ of the deepest convolution's accumulation terms for a $0.37$-point accuracy drop, and $25\%$ of the full-network arithmetic across the three deepest convolutions for a $1.36$-point drop, at a checkpoint cost negligible against the arithmetic removed. These figures are an arithmetic upper bound under an ideal per-unit execution model; making them hardware-realizable on larger binary backbones is the main direction ahead (Section~\ref{sec:disc-limits}).

\section*{Acknowledgment}

We would like to thank the Microelectronics and Microsystems Research Group (GRM) at Polytechnique Montréal for providing the computing infrastructure and GPU resources used in this work.

\bibliography{src/references}

@misc{iea2025energyai,
  author       = {{International Energy Agency}},
  title        = {Energy and {AI}},
  year         = {2025},
  address      = {Paris, France},
  howpublished = {\url{https://www.iea.org/reports/energy-and-ai}},
  note         = {Accessed: 2026-07-15}
}

@inproceedings{teerapittayanon2016branchynet,
  author    = {Teerapittayanon, Surat and McDanel, Bradley and Kung, H. T.},
  title     = {{BranchyNet}: Fast Inference via Early Exiting from Deep Neural Networks},
  booktitle = {Proceedings of the International Conference on Pattern Recognition},
  pages     = {2464--2469},
  year      = {2016}
}

@inproceedings{wang2018skipnet,
  author    = {Wang, Xin and Yu, Fisher and Dou, Zi-Yi and Darrell, Trevor and Gonzalez, Joseph E.},
  title     = {{SkipNet}: Learning Dynamic Routing in Convolutional Networks},
  booktitle = {Proceedings of the European Conference on Computer Vision},
  pages     = {409--424},
  year      = {2018}
}

@inproceedings{gao2019dynamicchannel,
  author    = {Gao, Xitong and Zhao, Yiren and Dudziak, Lukasz and Mullins, Robert and Xu, Cheng-Zhong},
  title     = {Dynamic Channel Pruning: Feature Boosting and Suppression},
  booktitle = {Proceedings of the International Conference on Learning Representations},
  year      = {2019}
}

@inproceedings{lecun1989optimal,
  author    = {LeCun, Yann and Denker, John S. and Solla, Sara A.},
  title     = {Optimal Brain Damage},
  booktitle = {Advances in Neural Information Processing Systems},
  volume    = {2},
  pages     = {598--605},
  year      = {1989}
}

@inproceedings{han2016deepcompression,
  author    = {Han, Song and Mao, Huizi and Dally, William J.},
  title     = {Deep Compression: Compressing Deep Neural Networks with Pruning, Trained Quantization and Huffman Coding},
  booktitle = {Proceedings of the International Conference on Learning Representations},
  year      = {2016}
}

@inproceedings{courbariaux2015binaryconnect,
  author    = {Courbariaux, Matthieu and Bengio, Yoshua and David, Jean-Pierre},
  title     = {{BinaryConnect}: Training Deep Neural Networks with Binary Weights during Propagations},
  booktitle = {Advances in Neural Information Processing Systems},
  volume    = {28},
  year      = {2015}
}

@inproceedings{courbariaux2016bnn,
  author    = {Courbariaux, Matthieu and Hubara, Itay and Soudry, Daniel and El-Yaniv, Ran and Bengio, Yoshua},
  title     = {Binarized Neural Networks: Training Deep Neural Networks with Weights and Activations Constrained to $+1$ or $-1$},
  booktitle = {Advances in Neural Information Processing Systems},
  volume    = {29},
  year      = {2016}
}

@inproceedings{rastegari2016xnor,
  author    = {Rastegari, Mohammad and Ordonez, Vicente and Redmon, Joseph and Farhadi, Ali},
  title     = {{XNOR-Net}: {ImageNet} Classification Using Binary Convolutional Neural Networks},
  booktitle = {Proceedings of the European Conference on Computer Vision},
  pages     = {525--542},
  year      = {2016}
}

@inproceedings{krizhevsky2012imagenet,
  author    = {Krizhevsky, Alex and Sutskever, Ilya and Hinton, Geoffrey E.},
  title     = {{ImageNet} Classification with Deep Convolutional Neural Networks},
  booktitle = {Advances in Neural Information Processing Systems},
  volume    = {25},
  year      = {2012}
}

@inproceedings{jouppi2017tpu,
  author    = {Jouppi, Norman P. and Young, Cliff and Patil, Nishant and Patterson, David and others},
  title     = {In-Datacenter Performance Analysis of a Tensor Processing Unit},
  booktitle = {Proceedings of the ACM/IEEE International Symposium on Computer Architecture},
  pages     = {1--12},
  year      = {2017}
}

@misc{mcdanel2017idp,
  author        = {McDanel, Bradley and Teerapittayanon, Surat and Kung, H. T.},
  title         = {Incomplete Dot Products for Dynamic Computation Scaling in Neural Network Inference},
  year          = {2017},
  eprint        = {1710.07830},
  archivePrefix = {arXiv},
  primaryClass  = {cs.LG},
  url           = {https://arxiv.org/abs/1710.07830}
}

@inproceedings{akhlaghi2018snapea,
  author    = {Akhlaghi, Vahid and Yazdanbakhsh, Amir and Samadi, Kambiz and Gupta, Rajesh K. and Esmaeilzadeh, Hadi},
  title     = {{SnaPEA}: Predictive Early Activation for Reducing Computation in Deep Convolutional Neural Networks},
  booktitle = {Proceedings of the 45th Annual International Symposium on Computer Architecture},
  pages     = {662--673},
  year      = {2018},
  doi       = {10.1109/ISCA.2018.00061}
}

@inproceedings{chen2019comprrae,
  author    = {Chen, Xi and Zhu, Jian and Jiang, Jingtong and Tsui, Chi-Ying},
  title     = {{CompRRAE}: {RRAM}-Based Convolutional Neural Network Accelerator with Reduced Computations through Runtime Activation Estimation},
  booktitle = {Proceedings of the 24th Asia and South Pacific Design Automation Conference},
  pages     = {133--139},
  year      = {2019},
  doi       = {10.1145/3287624.3287640}
}

@inproceedings{kong2023convrelu,
  author    = {Kong, Rui and Li, Yuesong and Yuan, Yun and Kong, Linghe},
  title     = {{ConvReLU++}: Reference-Based Lossless Acceleration of {Conv-ReLU} Operations on Mobile {CPU}},
  booktitle = {Proceedings of the 21st Annual International Conference on Mobile Systems, Applications, and Services},
  pages     = {503--515},
  year      = {2023},
  doi       = {10.1145/3581791.3596831}
}

@article{goldberg1991floating,
  author  = {Goldberg, David},
  title   = {What Every Computer Scientist Should Know about Floating-Point Arithmetic},
  journal = {ACM Computing Surveys},
  volume  = {23},
  number  = {1},
  pages   = {5--48},
  month   = mar,
  year    = {1991},
  doi     = {10.1145/103162.103163}
}

@techreport{krizhevsky2009cifar,
  author      = {Krizhevsky, Alex},
  title       = {Learning Multiple Layers of Features from Tiny Images},
  institution = {University of Toronto},
  year        = {2009}
}

\end{document}